\documentclass[10pt,twocolumn,letterpaper]{article}

\usepackage{iccv}
\usepackage{times}
\usepackage{epsfig}
\usepackage{graphicx}
\usepackage{amsmath}
\usepackage{amssymb}
\usepackage{indentfirst}
\usepackage[pagebackref=true,breaklinks=true,letterpaper=true,colorlinks,bookmarks=false]{hyperref}
\usepackage{multirow}
\usepackage{color, colortbl}
\definecolor{LightCyan}{rgb}{0.88,1,1}
\definecolor{Gray}{gray}{0.9}
\usepackage{makecell}
\usepackage{authblk}
\usepackage{hhline}
\usepackage{algorithm}
\usepackage{algpseudocode}
\usepackage{subcaption}
\usepackage{mwe}
\usepackage{tikz}

\newcommand*\circled[1]{\protect\tikz[baseline=(char.base)]{
            \protect\node[shape=circle,draw,inner sep=0.2pt] (char) {#1};}}
\makeatletter
\newcommand*{\bottop}{
  \mathpalette\@bottop{}
}
\newcommand*{\@bottop}[2]{
  \rlap{$#1\bot\m@th$}
  \top
}
\makeatother
\iccvfinalcopy
\def\iccvPaperID{11622}

\ificcvfinal\fi
\begin{document}

\title{ProtoFL: Unsupervised Federated Learning via Prototypical Distillation\vspace{-2.5ex}}

\author[1]{\emph{Hansol Kim}\thanks{\textbf{These authors contributed equally to this work}}}
\newcommand\CoAuthorMark{\footnotemark[\arabic{footnote}]}
\author[12]{\emph{Youngjun Kwak}\protect\CoAuthorMark}
\author[3]{\emph{Minyoung Jung}}
\author[1]{\emph{Jinho Shin}}
\author[4]{\emph{Youngsung Kim}}
\author[2]{\emph{Changick Kim\thanks{Corresponding author}}}

\affil[1]{KakaoBank Corp., South Korea}
\affil[2]{Department of Electrical Engineering, KAIST, South Korea}
\affil[3]{KETI, Korea Electronics Technology Institute, South Korea}
\affil[4]{Inha University, South Korea}
\affil[ ]{\small{\texttt{\{hans.kim,vivaan.yjkwak,william.shin\}@lab.kakaobank.com},
\texttt{\{yjk.kwak,changick\}@kaist.ac.kr}},
\texttt{minyoung.jung@keti.re.kr},
\texttt{y.kim@inha.ac.kr}}
\vspace{-3.5ex}

\maketitle
\ificcvfinal\thispagestyle{empty}\fi

\begin{abstract}
Federated learning (FL) is a promising approach for enhancing data privacy preservation, particularly for authentication systems. However, limited round communications, scarce representation, and scalability pose significant challenges to its deployment, hindering its full potential. In this paper, we propose `\textbf{ProtoFL}', \textbf{Proto}typical Representation Distillation based unsupervised \textbf{F}ederated \textbf{L}earning
to enhance the representation power of a global model and reduce round communication costs. Additionally, we introduce a local one-class classifier based on normalizing flows to improve performance with limited data. Our study represents the first investigation of using FL to improve one-class classification performance.
We conduct extensive experiments on five widely used benchmarks, namely MNIST, CIFAR-10, CIFAR-100, ImageNet-30, and Keystroke-Dynamics, to demonstrate the superior performance of our proposed framework over previous methods in the literature.
\end{abstract}

\begin{figure}[htb]
\centering
\includegraphics[width=\columnwidth]{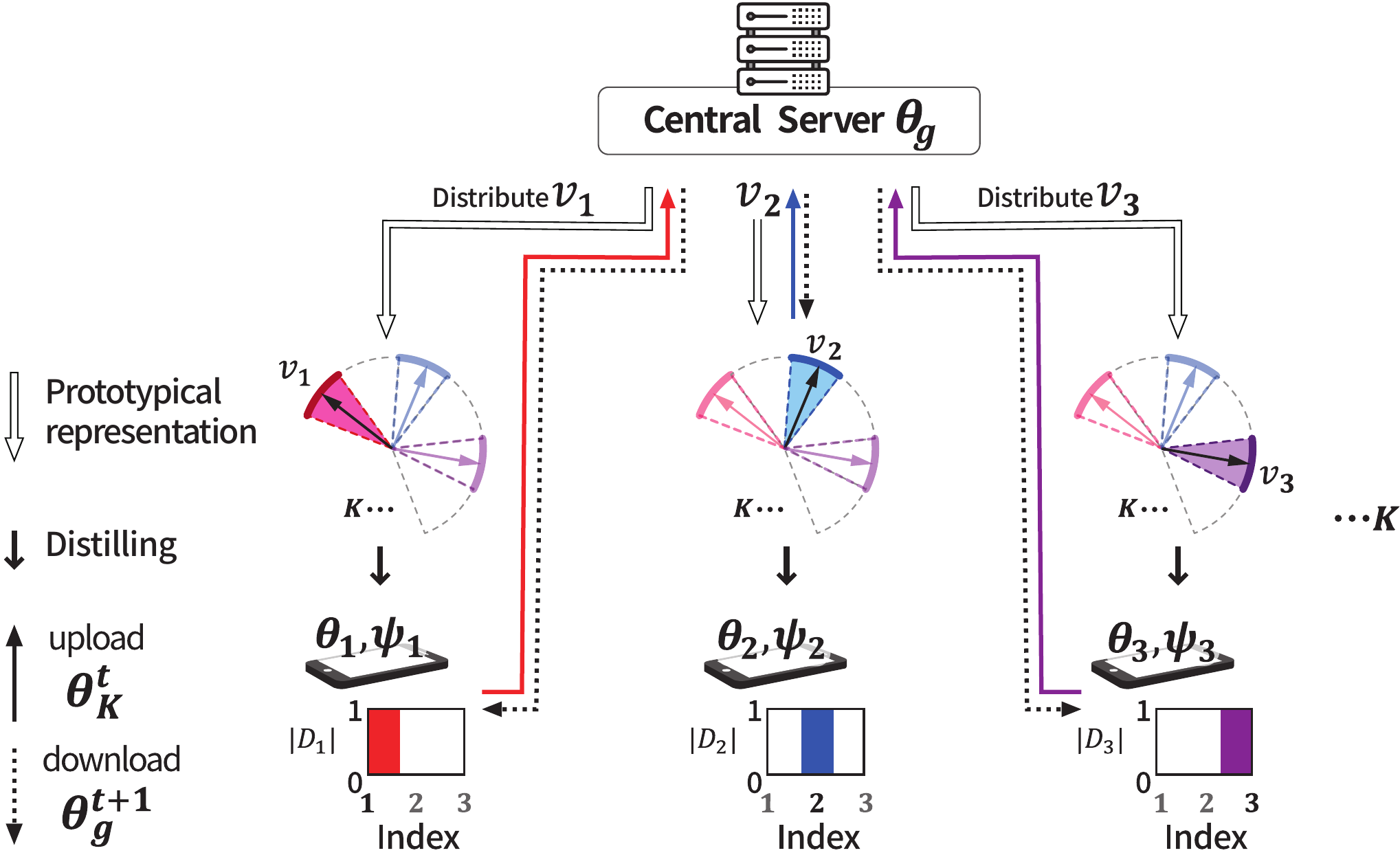}
\vspace{-1.2em}
\caption{Visualization of \textbf{extreme} non-i.i.d. ($\alpha=0.0$ is a concentration parameter of Dirichlet distribution) data based FL schema shows that each device has only the data from its target class not shared among all joined devices, and prototypical representation is distributed separately to corresponding joined clients in secure from a central server.}
\vspace{-0.5em}
\label{fig:fig_title}
\end{figure}

\vspace{-1.0em}
\section{Introduction}
In recent years, there has been a growing concern about privacy, leading people to hesitate when it comes to uploading their biological data to central data servers. Moreover, companies that possess personal information from users are strictly bound by the General Data Protection Regulation (GDPR)~\cite{truong2021privacy}. To address these privacy issues, Federated Learning (FL), an emerging distributed data parallel machine learning approach, has been proposed. FL leverages the decentralized data available on individual clients to collaboratively train a shared global model on a mediator server without the need to share personal data.

In the context of FL, where data is decentralized across individual clients, one-class classification (OCC) can be used. 
This is because that OCC can determine whether a new example belongs to the target distribution or not. Despite not using data from the non-target class, OCC has shown impressive performance~\cite{osti_6755553,Ruff2020Deep, tack2020csi, sohn2020learning, nfad, schmier2022anomaly,hendrycks2019oe,liznerski2022exposing}. 
In computer vision applications, OCC is particularly useful for detecting fraud, defects, and unauthorized users. 

Recent advancements in biometric authentication have highlighted the importance of FL-based OCC in computer vision (e.g., user-defined embedding-based methods~\cite{hosseini2021federated, fedfv} and data-driven methods~\cite{fedface,fedfr,fedaa}). 
However, FL-based OCC methods face significant challenges, including high communication costs, limited representation, and unstable learning processes. Additionally, centralized server-based methods~\cite{osti_6755553,sohn2020learning,tack2020csi} may not be suitable for FL-based OCC due to limited computing resources on each device, which are a major obstacle for centralized methods that require substantial computing power.

The limitations inherent in FL-based OCC necessitated the development of a novel approach to address these challenges. In this paper, we aim to attain a more expressive global model without the need for frequent re-training processes between the central server and client devices on a large scale. To achieve this, we propose \textbf{Proto}typical representation distillation based unsupervised \textbf{F}ederated \textbf{L}earning (ProtoFL) that distills representation from an off-the-shelf model learned using off-the-shelf datasets, regardless of individual client data. In contrast to traditional FL-based OCC, our proposed ProtoFL approach does not transfer parameters directly to the client, as depicted in Fig.~\ref{fig:fig_title}. This resolves the issues of frequent round communication costs and the need for extra data due to one-time prototype representation distillation. ProtoFL provides a novel solution to the existing challenges of FL-based OCC, enabling efficient and effective global model updates. 

Subsequently, we suggest a novel approach for estimating the density of a target class in a distributed setting using a flow-based one-class classifier~\cite{NEURIPS2020_ecb9fe2f}. To achieve this, we conduct the estimation on individual client devices, using augmented latent variables for training their distributed models. Our approach leverages two key techniques: maximum likelihood estimation with $\log$-likelihood and a probabilistic similarity loss function that includes $\mathcal{KL}$-divergence. By combining the distillation and one-class classification phases, we can effectively handle complex data distributions that are non-independent and non-identically distributed across individual clients. Our two-phase learning framework is inspired by the success of flow-based models in various applications~\cite{NEURIPS2020_ecb9fe2f,gudovskiy2022cflow, schmier2022anomaly, CHO2022108703}, and we demonstrate its effectiveness in handling complex data distributions in the distributed settings.  

The experimental findings of our proposed method demonstrate superior classification performance compared to both server-based and client-based methods on both image and tabular datasets. Additionally, we have validated the scalability of the learned representation and have shown that the global model learned by the ProtoFL is compatible with existing one-class classifiers as well as our one-class classifier based on the benchmark datasets. Our results indicate that our method is a promising approach for large-scale machine learning tasks that require robust and scalable classification capabilities.

Our contributions are described as follows:
\begin{itemize}
    \item We propose a novel unsupervised federated learning framework that effectively addresses the challenge of insufficient local training data. By leveraging normalizing flows for local classifier learning and prototypical representation distillation, our approach enables efficient and effective global model updates.   \item We propose a novel prototype-based representation learning method for distilling normal data representation using an off-the-shelf model and dataset. Our approach demonstrates the scalability of the global model, which can be verified by adding new clients in FL-based OCC.
    \item We propose new federated and centralized learning methods for one-class classification, which we evaluate on five widely-used benchmarks. Our experiments show that our methods achieve superior performance compared to existing approaches.
\end{itemize}

\vspace{-1.0em}
\section{Related Work}
\paragraph{One-class classification}
Various one-class classification approaches have been proposed and categorized into description-based and representation-based learning. In the description-based methods, Deep-SVDD~\cite{osti_6755553} performs one-class detection by learning a model to map target samples into a center in the latent space, otherwise non-target samples are mapped far from the center. And FCDD~\cite{liznerski2021explainable} proposes an explainable one-class classifier by upsampling based on gaussian sampling. In contrast, representation-based learning methods (DROC~\cite{sohn2020learning} and CSI~\cite{tack2020csi}) present a two-step learning framework to learn a representation model through excessive data-augmentation and contrastive loss. The framework either employs a classifier or defines a score function for detecting a target class. Unlike previous centralized server-based approaches, our proposed approach for one-class classification on decentralized learning avoids the risk of personal data leakage by constructing a model using distributed data. To the best of our knowledge, our method is the first to directly address this problem in a decentralized setting.

\vspace{-0.5em}
\paragraph{Federated learning for local one-class classifier}
Federated Learning (FL) is a distributed machine learning paradigm that enables collaborative model training without direct data sharing. FL has been applied to one class classification (OCC) tasks for user verification and authentication, with several studies demonstrating promising results.~\cite{hosseini2021federated, fedaa, parkfederated}.
FedAwS~\cite{yu2020federated} introduces a geometric regularization to learn a global model by utilizing uploaded latent variables of joined clients in a central server. Since the latent variables shared on a server are private data, the FedAwS violates in the setting of FL. FedUV~\cite{hosseini2021federated} employs independent secret error correcting codes (ECCs) to train a one-class classifier by preventing personal data from sharing among joined clients. The secret codes induce the concise objective of FedUV so that only positive examples are required. The FedUV approach was further refined to estimate a center of distribution by FedAA~\cite{fedaa} and FedMetric~\cite{parkfederated}, instead of defining hand-designed codes. 
However, local data on each device is not sufficient to represent a centroid of local data distribution. 

\vspace{-0.5em}
\paragraph{Federated learning for central and client classifiers}
In authentication systems, client data is often partitioned according to a target class. This results in highly or extremely non-i.i.d. data that poses a significant challenge for FL. This \textbf{extreme} non-i.i.d. means that a concentration parameter of Dirichlet distribution approaches to zero ($\alpha=0.0$) while many federated learning methods~\cite{collins2021exploiting, SphereFed, han2022fedx} assumes non-i.i.d. data with non-zero ($\alpha~\textcolor{red}{\neq}~0.0$) setting~\cite{noniid,liang2020think, Orchestra,collins2021exploiting,han2022fedx,SphereFed}.

Unsupervised federated learning methods~\cite{liang2020think, Orchestra} aggregate local representation models and centroids for achieving a global model and centroid. FedRep~\cite{collins2021exploiting} shares partitions of a global model to adapt each local heterogeneous data. FedX~\cite{han2022fedx}, which utilizes structure knowledge distillation between local and global knowledge relationships, learns meaningful data representation without sharing external data. SphereFed~\cite{SphereFed} introduced a learned matrix, which is a fixed-classifier for sharing among all participating clients, to transform each local data distribution into the predefined latent space. But, SphereFed requires enormous cost to re-construct and re-distribute a new learned matrix whenever a new client joins. 
Therefore, we first propose a prototypical representation distillation learning to save communication cost in FL with \textbf{extreme} non-i.i.d. ($\alpha=0.0$) data settings.

\vspace{-0.5em}
\paragraph{Normalizing flows as classifiers}
Normalizing flows (NFs) normalize entangled data distributions into disentangled distributions by composing invertible and differentiable transformations. NFs have been applied for AD in image and video tasks~\cite{gudovskiy2022cflow, schmier2022anomaly, CHO2022108703} because FOOD~\cite{NEURIPS2020_ecb9fe2f} finds that NFs transforming latent to latent space outperforms those transforming data to latent space. For instance, FLOW~\cite{schmier2022anomaly} utilizes a fixed feature extractor to train NFs by either maximizing the $\log$-likelihood on a target-class or minimizing the $\log$-likelihood on outlier-exposure (OE) data. ITAE~\cite{CHO2022108703} employs NFs to estimate the density by learning appearance and motion latent features in videos. By the effectiveness of AD using NFs, we apply the NFs minimizing $\log$-likelihood to our global model for one-class classification.

\section{Preliminary}

\paragraph{Federated Average Learning}
FedAVG~\cite{mcmahan2017communication}, composed of participating clients and a central mediator server, trains a global model by
aggregating locally-computed parameters, and broadcasts the updated global model to the clients. In each round, FedAVG updates the global model parameters with local model parameters as follows:
\begin{align}
    \theta_{g}^{t+1} = \sum_{k=1}^K \frac{|D_{k}|}{\sum_{k=1}^{K}|D_k|} \theta_k^t,
\label{eq1:fedavg}
\end{align}
where $k \in \{1, ..., K\}$ indicates an index of a local client. $\theta_{g}^{t+1}$ is the parameters of a global model, and $\theta_k^t$ is the parameters of the $k^{th}$ local model at a round $t$. For the balanced updating of the global model $\theta_{g}^{t+1}$, $|D_k|$ is the number of samples on dataset $D_k$.

\vspace{-0.5em}
\paragraph{Unsupervised Contrastive Learning}
FL with \textbf{extreme} non-i.i.d.($\alpha=0.0$) data setting is unavailable to access samples of the other clients. Thus, we simplify the contrastive loss, $\mathcal{L}_{cntr} = y\times (1-d(\bullet,\hat{\bullet}))
    + (1-y)\times \max (0, d(\bullet,\hat{\bullet}))$, as follows:

\vspace{-0.5em}
\begin{align}
    \mathcal{L}_{d} = 1-d(\bullet,\hat{\bullet}),
\label{eq2:lp}
\end{align}
where $d(\bullet,\hat{\bullet})$ is the cosine-similarity $\frac{\bullet \cdot \hat{\bullet}}{\|\bullet\|\cdot\|\hat{\bullet}\|}$. We utilize $L_{d}$ as our positive cosine similarity loss for our unsupervised leaning, instead of $\mathcal{L}_{cntr}$.

\begin{figure*}[htb]
\includegraphics[width=\textwidth]{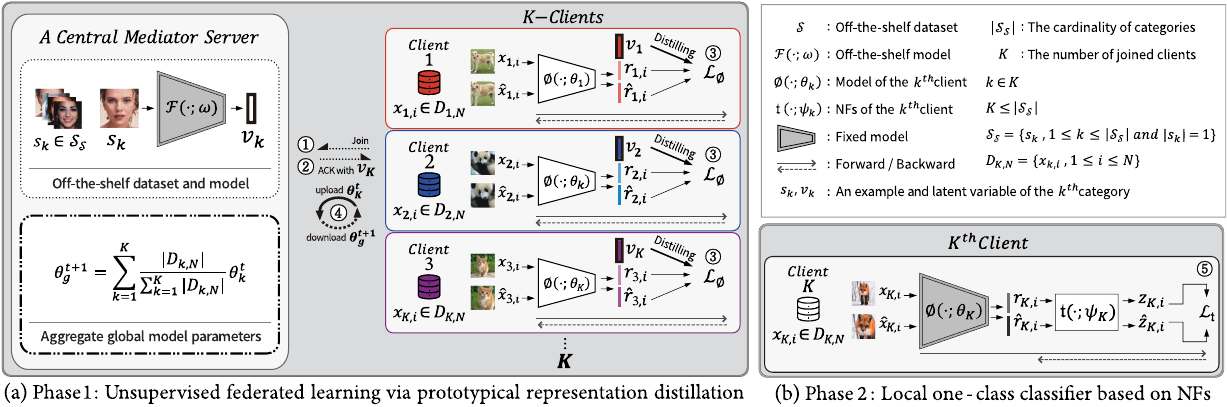}\\
\vspace{-1.5em}
\caption{Overview of our proposed two-phase learning architecture. \circled{1} $\sim$ \circled{4} indicates the workflows of ProtoFL with all the joined clients to upload parameters of each client and to download the aggregated global model, and \circled{5} represents a local training for the OC-NF. The details of sequences \circled{1} $\sim$ \circled{5} are described in Section ~\ref{procedures}. NFs denote normalizing flows.}
\vspace{-1.0em}
\label{fig:fig_main}
\end{figure*}

\vspace{-0.5em}
\paragraph{Normalizing Flows}
Normalizing flows (NFs)~\cite{dinh2017density} are the statistical methods using the change-of-variable law of probabilities to fit an arbitrary target density $p_{R}(r)$ by a tractable base distribution with density $p_{Z}(z)$ and a bijective invertible mapping $\mathfrak{t}^{-1}:R^{D} \rightarrow Z^{D} \Leftrightarrow \mathfrak{t}:Z^{D} \rightarrow R^{D}$. According to ~\cite{JMLR:v22:19-1028}, $\mathcal{KL}$-divergence and $\log$-likelihood estimation are employed to optimize the invertible flow-based model $z=\mathfrak{t}^{-1}(r;\psi)$ and $r=\mathfrak{t}(z;\psi)$ by the target distribution $p_{R}(r)$. The bijective invertible mapping is drawn as:

\vspace{-1.5em}
\begin{equation}
\begin{aligned}
    &\mathcal{D}_{\mathcal{KL}}[\hat{p}_{R}(r)\|p_{R}(r;\psi)] \\
    &\approx -\mathbb{E}_{\hat{p}_{R}(r)}[\log p_{Z}(\mathfrak{t}^{-1}(r;\psi)) + \log |\det\mathfrak{j}_{\mathfrak{t}^{-1}}|] + \textit{cont},
\end{aligned}
\label{eq3:ocnf}
\end{equation}
where $p_{R}(r;\psi)$ denotes the likelihood of a model. $\hat{p}_{R}$ is the target distribution to learn the model by $\mathcal{KL}$-divergence. $p_{R}(r;\psi)$ only remains for learning parameters $\psi$, and then is replaced by $\frac{1}{N} \sum^{N}_{i=1}\left[ \frac{\|\mathfrak{t}^{-1}(r_{i})\|^{2}_{2}}{2} - \log |\det \mathfrak{j}_{\mathfrak{t}^{-1}}| \right]$ through the training dataset $D_{k}$ as explained by ~\cite{gudovskiy2022cflow, NEURIPS2020_ecb9fe2f}. The re-written equation is applied to our objective to estimate the $\log$-likelihood of given local data on each device.

\section{Proposed Method}
In this section, we propose a two-phase unsupervised learning framework that combines unsupervised federated learning via distillation of prototypical representations with local classifier learning via leveraging normalizing flows. 

\vspace{-0.5em}
\subsection{Problem Formulation}
Participating $K$ clients aim to train a global model by aggregating locally-computed parameters $\phi(\bullet,\theta)$ for an one-class classification task without sharing private data among joined clients.
Each client has a training dataset $D_{K,N} =\{x_{k,i}, 1 \leq i \leq N\ , 1 \leq k \leq K\}$ and $N$ is the cardinality of the local data $D_{k,N}$. Let $\phi(x_{k,i}; \theta_{k})$ denote the trainable model of the $k^{th}$ client.
In particular, we follow the \textbf{extreme} non-i.i.d. ($\forall_{m,n} \ D_{m,N} \cap D_{n,N} = \emptyset$) data setting in our FL-based OCC. Different from prior FL settings utilizing either hand-designed codes or additional training datasets, we invent generic representation of the local model $\phi$ by leveraging the off-the-shelf model $\mathcal{F}$ and dataset $\mathcal{S}$, and we employ the flow-based model $\psi$ for the local one-class classifier.

\vspace{-0.5em}
\subsection{Sequence of procedures}\label{procedures}
As shown in Fig.~\ref{fig:fig_main}, we describe our proposed method workflow in each phase.
\circled{1} Whenever a new client participates in our unsupervised federated learning, all clients join a central mediator server. \circled{2} For the $k^{th}$ client, a categorical image $s_{k}$ is chosen from $\mathcal{S}$ and transformed into a latent variable $v_{k}$ by the central mediator server. Both \circled{1} and \circled{2} occur only once at the first time, and $\mathcal{V}=\{v_{k}, 1 \leq k \leq K\}$ is secretly distributed to the corresponding client. \circled{3} We train the $k^{th}$ local model $\theta_{k}$ via the local data $D_{k,N}$, the distributed prototype representation $v_{k}$, cosine similarly loss, and $\mathcal{KL}$-divergence loss. \circled{4}
A global model $\theta_{g}$ is aggregated with all the uploaded local models by FedAVG, and the global model is distributed to all participating clients. \circled{5} Once federated representation learning has finished, we train a local one-class classifier based on normalizing flows (OC-NF) by using maximum likelihood and cosine similarity losses. We describe the details of our procedures(\circled{1} $\sim$ \circled{5}) in the following subsections.

\subsection{Prototypical representation distillation based unsupervised federated learning (ProtoFL)}
\begin{figure}[tb]
\includegraphics[width=\columnwidth]{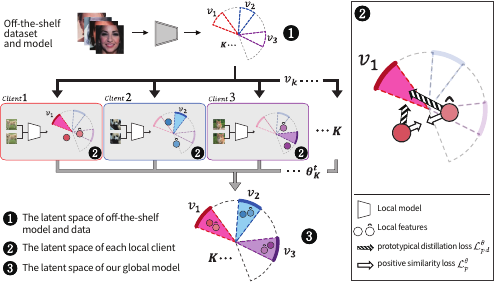}\\
\vspace{-2.0em}
\caption{Illustration of learning a global model with prototypical representation distillation on distributed devices without sharing private data among participating clients.}
\vspace{-1.0em}
\label{fig:fig_compare}
\end{figure}

We assume $\mathcal{F}: R^{W \times H \times C} \rightarrow R ^{D}$ and $\mathcal{\phi}: R^{W \times H \times C} \rightarrow R ^{D}$ indicate the off-the-shelf model and the local model, respectively as illustrated in Fig.~\ref{fig:fig_main}. For training the local model of the $k^{th}$ client, we augment two views $x_{k, i}$, $\hat{x}_{k, i}$ from the $i^{th}$ example of the local training data $D_{k,N}$, and subsequently transform the two samples as follows:

\vspace{-1.0em}
\begin{align}
    r_{k, i} =\mathcal{\phi}(x_{k, i};\theta_{k});~
    \hat{r}_{k, i} =\mathcal{\phi}(\hat{x}_{k, i};\theta_{k}),
\end{align}
where $r_{k, i}$ and $\hat{r}_{k, i}$ represent latent variables of the $i^{th}$ example. $\mathcal{\phi}$ is the client model with $k^{th}$ learnable parameters $\theta_{k}$. Inspired by SimCLR~\cite{chen2020simple}, the two latent variables $r_{k}$ and $\hat{r}_{k}$ must be very close to each other. Given the only positive local data and the constraint, we employ the positive cosine similarity loss(Eq.~\ref{eq2:lp}) to learn the local model as follows:

\vspace{-1.0em}
\begin{align}
    \mathcal{L}^{\theta}_{p}(\theta_{k}) = \frac{1}{N} \sum^{N}_{i=1} \left[ 1- d(r_{k, i} ,\hat{r}_{k, i}) \right],
\label{eq5:positive_cosine_similarity}
\end{align}
where $\theta_{k}$ is optimized by the local data $D_{k,N}$. To overcome the small cardinality of local data for training the local representative model, we propose the off-the-shelf model with pre-trained parameters $\mathcal{F}({\bullet; \omega)}$ and the off-the-shelf dataset $s_{k} \in \mathcal{S}_{\mathfrak{s}}$. The off-the-shelf model and dataset are utilized on a central mediator server as depicted:
\begin{equation}
     v_{k} = \mathcal{F}(s_{k};\omega),
\end{equation}
where $s_{k}$ and $v_{k}$ represent a prototype example and representation for the $k^{th}$ client. After $\mathcal{V}=\{v_{k}, 1 \leq k \leq K\}$ is distributed to the corresponding client in secret, we train the local model of the $k^{th}$ client to estimate the prototypical target representation $v_{k}$. We employ the $\mathcal{KL}$-divergence to minimize the discrepancy between two distributions $v_{k}$ and either $r_{k}$ or $\hat{r}_{k}$. Therefore, we invent the prototypical distillation loss as follows:

\vspace{-1.0em}
\begin{align}
    \mathcal{L}^{\theta}_{pd}(\theta_{k}) &= \frac{1}{N} \sum^{N}_{i=1} \left[ \mathcal{KL}(v_{k} \| r_{k,i}) + \mathcal{KL}(v_{k} \| \hat{r}_{k,i}) \right] ,
\label{eq7:kl_divergence}
\end{align}
where $\mathcal{KL}(v_{k} \| r_{k,i})$ and $\mathcal{KL}(v_{k} \| \hat{r}_{k,i})$ are equally contributed to our distillation loss. Figure~\ref{fig:fig_compare} describes the details of Eq.~\ref{eq5:positive_cosine_similarity} and Eq.~\ref{eq7:kl_divergence}, and the overall objective of the first phase (phase 1) is defined as follows:
\begin{align}
    \mathcal{L}_{\phi}(\theta_{k})  =(1-\alpha) \times  \mathcal{L}^{\theta}_{pd}(\theta_{k})  + \alpha \times \mathcal{L}^{\theta}_{p}(\theta_{k}) ,
\label{eq8:representation_loss}
\end{align}
where $\alpha$ is a balancing weight. In our experiment, $\alpha$ is empirically set to $0.1$.

\subsection{Local one-class classifier via Normalizing Flows (OC-NF)}
After unsupervised federated representation learning for the global model, we leverage a general flow-based model for training the local one-class classifier in each client as shown in Fig.~\ref{fig:fig_main}. To estimate $\log$-likelihood of transformed latent variables $r_{k}$ and $\hat{r}_{k}$, we exploit Eq.~\ref{eq3:ocnf} to learn the probabilistic normalizing flows $\mathfrak{t}$ through the local dataset $D_{k,N}$ and the local model $\phi(\cdot;\theta_{k})$ on the $k^{th}$ client as follows:

\vspace{-1.0em}
\begin{align}
 & \mathcal{L}^{\psi}_{mle}(\psi_{k}) \nonumber\\ 
= & \frac{1}{N} \sum^{N}_{i=1} 
\left[ \frac{\|\mathfrak{t}^{-1}(r_{k,i};\psi_{k})\|^{2}_{2}}{2} - \log |\det  
\mathfrak{j}_{k,i}| \right] + \textit{cont} \nonumber \\
\approx& \frac{1}{N} \sum^{N}_{i=1} 
\left[ \frac{\|z_{k,i}\|^{2}_{2}}{2} - \log |\det  
\mathfrak{j}_{k,i}| \right] + \textit{cont},
\label{eq9:maximize_loglikelihood}
\end{align}
where $z_{k,i}$ is the estimated distribution as it passes through $\mathfrak{t}^{-1}(r_{k,i};\psi_{k})$. $r_{k,i}$ is a latent variable from $\phi(x_{k,i};\theta)$. When optimizing the flow-based model, we ignore constants, $\det$ and $\textit{cont}$. Since $r_{k,i}$ and $\hat{r}_{k,i}$ are close each other in a latent space, $z_{k,i}$ and $\hat{z}_{k,i}$ also predict the same distribution so that we introduce the following objective as a regularizer:
\begin{align}
    \mathcal{L}^{\psi}_{reg}(\psi_{k})= \frac{1}{N} \sum^{N}_{i=1} \left[ 1- d(z_{k, i} ,\hat{z}_{k, i}) \right],
\label{eq10:positive_cosine_similarity}
\end{align}
where $z_{k,i}$ and $\hat{z}_{k,i}$ are the normal distribution estimated by $\mathfrak{t}^{-1}((\phi(x_{k,i};\theta_{k});\psi_{k}))$ and $\mathfrak{t}^{-1}((\phi(\hat{x}_{k,i};\theta_{k});\psi_{k}))$ respectively. The overall objective of the second phase (phase 2) is composed of Eq. \ref{eq9:maximize_loglikelihood} and Eq. \ref{eq10:positive_cosine_similarity} as follows:
\begin{align}
    \mathcal{L}_{\mathfrak{t}}(\psi_{k})= \mathcal{L}^{\psi}_{mle}(\psi_{k})+ \lambda \times  \mathcal{L}^{\psi}_{reg}(\psi_{k}),
\label{eq11:local_training_loss}
\end{align}
where $\lambda$ is a hyper-parameter for effectiveness of regularization. For our experiment, we empirically set $\lambda$ to $0.01$.

\section{Experiments}
In this section, we demonstrate the proposed approach on image and tabular benchmarks. We evaluate the performance of our proposed ProtoFL and compare our model with the other FL methods and one-class detectors. And we analyze the impact of new client participation in the context of OCC on FL, as well as ablation studies on each proposed component.

\subsection{Datasets}
We thoroughly evaluate our proposed method on several benchmark datasets, including the widely used MNIST~\cite{lecun2010mnist}, CIFAR-10~\cite{krizhevsky2009learning}, CIFAR-100~\cite{krizhevsky2009learning}, ImageNet-30~\cite{hendrycks2019using}, and Keystroke-Dynamics~\cite{killourhy2009comparing}, to tackle the demanding task of one-class classification. Our evaluation scrutinizes the efficacy, robustness, and scalability of our method, highlighting its strengths in handling real-world scenarios. In case of CIFAR-100, we adopt 20 super-class labels, denoted by CIFAR-100\textsuperscript{\ddag}, as suggested in ~\cite{golan2018deep} to our experiments. The numbers of categories in the benchmarks are 10, 10, 20, 30, and 51, which are the numbers of participating clients on our federated learning. 
The vision and tabular based benchmarks for OCC follow a one-vs-rest protocol~\cite{golan2018deep, sohn2020learning, tack2020csi}. In the protocol, a set of samples from one class indicates a target class for one client, whereas a set of samples from the remaining classes represents a non-target class for the remaining clients. 

\subsection{Experiment Setting}
\paragraph{Off-the-shelf model and dataset}
In the central mediator server, the off-the-shelf dataset and model generate prototypical representation when a new client participates in our unsupervised federated learning. For the off-the-shelf model and dataset, we exploit ArcFace~\cite{Deng_2019_CVPR} based ResNet-50 backbone~\cite{he2016deep} and MS-Celeb-1M~\cite{guo2016ms}, respectively.

\vspace{-1.0em}
\paragraph{Architectures}
In this experiment, we employ ResNet-18/32 backbones~\cite{he2016deep} to learn the local and global models on our proposed approach. Since batch normalization is detrimental to the performance in federated learning~\cite{hosseini2021federated} both ResNet-18/32 backbones replace batch- with group-normalization. In the Keystroke Dynamics dataset, we employ multi-layer perceptions as the local and global model, consisting of three linear layers, ReLU, and group-normalization. And we utilize NFs as a local classifier model composed of 8 coupling layers~\cite{dinh2017density}.

\begin{algorithm}[tb]
\caption{\small Procedure of our two-phase learning}\label{alg:cap1}
\begin{algorithmic}
\small
\State \emph{\textbf{Phase 1 (ProtoFL) :} Start phase 1 for T rounds.}
% \Indent
\State Initialize and broadcast a global model $\theta_g^0$ to all joined $K$ clients. Send an acknowledgement (ACK) with prototypical representation $v_{k}$ to the $k^{th}$ client.
\For {$t \gets 0$ to $T-1$}
    \State $\theta_g^{t+1} \gets$ GLOBAL\_TRAINING($t, \theta_g^t$)
\EndFor
\State Update the global model $\theta_g$ with $\theta_g^T$.
% \EndIndent

\Function{GLOBAL\_TRAINING}{$t, \theta_g^t$}
    \State Randomly select $k$ number of clients from $\{C_k\}_{k=1}^K$.
    \For{$k \gets$ 1 to $K$}
        \State Broadcast the global model $\theta_g^t$ to $C_k$.
        \State $\theta_k^t, |D_{k,N}| \gets$ CLIENT\_TRAINING($t, k, \theta_g^t$)
    \EndFor
    \State $\theta_g^{t+1} \gets \sum_{k=1}^{K}\frac{|D_{k,N}|}{\sum_{k=1}^K|D_{k,N}|}\theta_k^t$
    \State \Return $\theta_g^{t+1}$
\EndFunction
\Function{CLIENT\_TRAINING}{$t, k, \theta_g^t$}
    \State Download the global model $\theta_g^t$, and assign it to $\theta_k^t$.
    \For{$x_{k,i} \in D_{K,N}$}
        \State $\mathcal{L}_{\phi}(\theta_k^t) = (1-\alpha) \times  \mathcal{L}_{pd}^{\theta}(\theta_k^t)  + \alpha \times \mathcal{L}_{p}^{\theta}(\theta_k^t)$
        \State $\theta_k^t \gets \theta_k^t - \eta\nabla \mathcal{L}_{\phi}(\theta_k^t)$
    \EndFor
    \State Calculate the number of the $k^{th}$ client data $D_{k,N}$.
    \State \Return $\theta_k^t$ and $|D_{k,N}|$
\EndFunction
\State \emph{\textbf{Phase 2 (OC-NF) : } All joined clients start Algorithm~\ref{alg:cap2} for OCC.}
\end{algorithmic}
\end{algorithm}

\begin{algorithm}[tb]
\caption{\small Procedure of our OC-NF for OCC}\label{alg:cap2}
\begin{algorithmic}
\small
\State Download the global model $\theta_g$, and assign it to the $k^{th}$ client model $\theta_k$. Freeze the global representation model $\phi(\bullet;{\theta_k})$. Initialize the flow-based model of the $k^{th}$ client $\mathfrak{t}_{\psi_{k}}$.
\For{$x_{k,i} \in D_{k,N}$}
    \State $\mathcal{L}_{\mathfrak{t}}(\psi_{k})=   \mathcal{L}^{\psi}_{mle}(\psi_{k})+ \lambda \times  \mathcal{L}^{\psi}_{reg}(\psi_{k})$
    \State $\psi_{k} \gets \psi_{k} - \eta\nabla \mathcal{L}_{\mathfrak{t}}(\psi_{k})$
\EndFor
\end{algorithmic}
\end{algorithm}

\vspace{-0.3em}
\paragraph{Implementation details}
In the first phase as described in Algorithm~\ref{alg:cap1}, we train our expressive global model based on the FedAvg~\cite{mcmahan2017communication} method with 1 local epoch and 900 communication rounds, using randomly chosen clients from all participating clients. We utilize the RAdam optimizer~\cite{liu2019variance} with betas 0.94 and 0.98, a weight decay 1e-3 and a learning rate 1e-6 to learn local models. In the second phase as described in Algorithm~\ref{alg:cap2}, we use the SGD optimizer~\cite{robbins1951stochastic} with a learning rate 5e-3 and 5 epochs to learn our one-class classifier. Data augmentations include the set of random processes (crop, horizontal-flip, and gaussian-blur) and color jitter. To evaluate the performance of various methods, we use the area under curve of ROC curve (AUROC) and equal error rate (EER).

\begin{table}[h]
\begin{center}
\resizebox{\columnwidth}{!}{
\begin{tabular}{|l|c|cc|cc|}
\hline
\multicolumn{1}{|c|}{\multirow{2}{*}{Method}} & \multirow{2}{*}{Network} & \multicolumn{2}{c|}{MNIST} & \multicolumn{2}{c|}{CIFAR-10} \\ \cline{3-6} 
\multicolumn{1}{|c|}{} &  & \multicolumn{1}{c|}{AUROC} & Round & \multicolumn{1}{c|}{AUROC} & Round \\ \hline
FedAwS~\cite{yu2020federated} & ResNet-32 & \multicolumn{1}{c|}{99.6} & 10K & \multicolumn{1}{c|}{94.1} & 100K \\ \hline
FedUV~\cite{hosseini2021federated} & ResNet-32 & \multicolumn{1}{c|}{99.7} & 20K & \multicolumn{1}{c|}{87.2} & 20K \\ \hline
FedMetric~\cite{parkfederated} & ResNet-32 & \multicolumn{1}{c|}{99.6} & 10K & \multicolumn{1}{c|}{94.2} & 100K \\
 \hline
\rowcolor{Gray} Ours & ResNet-32 & \multicolumn{1}{c|}{\textbf{99.9}} & \textbf{0.9K} & \multicolumn{1}{c|}{\textbf{95.2}} & \textbf{0.9K} \\ \hline
\end{tabular}
}
\end{center}
\vspace{-2.0em}
\caption{Performance comparison FL-based methods with our proposed approach on MNIST and CIFAR-10 benchmarks. Our approach outperforms both benchmarks while reducing the communication round cost.}
\label{tab:table1}
\end{table}
\vspace{-1.0em}

\begin{table}[h]
\centering
\resizebox{\columnwidth}{!}{
\begin{tabular}{|c|l|c|c|c|c|}
\hline
\multirow{2}{*}{Type} & \multicolumn{1}{c|}{\multirow{2}{*}{Method}} & \multirow{2}{*}{Network} & CIFAR-10 & CIFAR-100\textsuperscript{\ddag} & ImageNet-30 \\ \cline{4-6} 
 & \multicolumn{1}{c|}{} &  & AUROC & AUROC & AUROC \\ \hline
\multirow{8}{*}{\begin{tabular}[c]{@{}c@{}}Centralized \\ Learning\end{tabular}} & AE~\cite{liznerski2021explainable} & - & - & - & 56.0 \\ \cline{2-6} 
 & OC-SVM~\cite{scholkopf1999support} & - & 58.8 & 63.1 & - \\ \cline{2-6} 
 & Geom~\cite{golan2018deep} & WRN-16-8 & 86.0 & 78.7 & - \\ \cline{2-6} 
 & Rot~\cite{hendrycks2019using} & ResNet-18 & 89.8 & 77.7 & 77.9 \\ \cline{2-6} 
 & GOAD~~\cite{bergman2020classification} & ResNet-18 & 85.1 & 74.5 & - \\ \cline{2-6} 
 & DROC~\cite{sohn2020learning} & ResNet-18 & 92.5 & 86.5 & - \\ \cline{2-6} 
 & CSI~~\cite{tack2020csi} & ResNet-18 & 94.3 & 86.6 & 91.6 \\ \cline{2-6} 
 & FLOW$^{\dag}$~\cite{schmier2022anomaly} & ResNet-50 & \underline{95.2} & \textbf{93.0} & - \\ \hline
\multirow{3}{*}{\begin{tabular}[c]{@{}c@{}}Federated \\ Learning\end{tabular}} & FedUV~\cite{hosseini2021federated} & ResNet-18 & 79.8\textsuperscript{*} & 55.9\textsuperscript{*} & 62.8\textsuperscript{*} \\ \cline{2-6} 
 & FedRep~\cite{collins2021exploiting} & ResNet-18 & 58.2\textsuperscript{*} & 56.9\textsuperscript{*} & 56.3\textsuperscript{*} \\ \cline{2-6} 
 & \cellcolor[gray]{0.9}Ours & \cellcolor[gray]{0.9}ResNet-18 & \cellcolor[gray]{0.9}\textbf{95.3} & \cellcolor[gray]{0.9}\underline{89.9} & \cellcolor[gray]{0.9}\textbf{95.4} \\ \hline
\end{tabular}
}
\vspace{-1.0em}
\caption{AUROC of various centralized and decentralized detection methods on CIFAR-10/100\textsuperscript{\ddag}, and ImageNet-30 for OCC. $^{\dag}$ and \textsuperscript{*} denote the usage of a pre-trained model learned on ImageNet-1M and the values from our re-implementation respectively, whereas \textbf{bold} and \underline{underline} indicate the best results and the second results, respectively.}
\vspace{-1.0em}
\label{tab:table2} 
\end{table}

\subsection{Experimental Results}

\begin{table*}[]
\begin{center}
\resizebox{\textwidth}{!}{
\begin{tabular}{|c|l|l|cccccccccc||c|}
\hline
Type & \multicolumn{1}{c|}{Method} & \multicolumn{1}{c|}{Network} & Plane & Car & Bird & Cat & Deer & Dog & Forg & Horse & Ship & Truck & \multicolumn{1}{c|}{Mean (std)} \\ \hline
\multirow{11}{*}{\begin{tabular}[c]{@{}c@{}}Centralized \\ Learning\end{tabular}} 
 & OC-SVM~\cite{scholkopf1999support} & - & 65.6 & 40.9 & 65.3 & 50.1 & 75.2 & 51.2 & 71.8 & 51.2 & 67.9 & 48.5 & 58.8 (±11.6) \\ \cline{2-3} 
 & DeepSVDD~\cite{ruff2018deep} & LeNet & 61.7 & 65.9 & 50.8 & 59.1 & 60.9 & 65.7 & 67.7 & 67.3 & 75.9 & 73.1 & 64.8 (±7.17) \\ \cline{2-3}
 & Geom~\cite{golan2018deep} & WRN-16-8 & 74.7 & 95.7 & 78.1 & 72.4 & 87.8 & 87.8 & 83.4 & 95.5 & 93.3 & 91.3 & 86.0 (±8.52) \\ \cline{2-3} 
 & Rot~\cite{hendrycks2019using} & WRN-16-4 & 77.5 & 96.9 & 87.3 & 80.9 & 92.7 & 90.2 & 90.9 & 96.5 & 95.2 & 93.3 & 90.1 (±6.52) \\ \cline{2-3} 
 & GOAD~\cite{bergman2020classification} & WRN-16-4 & 77.2 & 96.7 & 83.3 & 77.7 & 87.8 & 87.8 & 90.0 & 96.1 & 93.8 & 92.0 & 88.2 (±6.99) \\ \cline{2-3} 
 & Rot~\cite{hendrycks2019using} & ResNet-18 & 80.4 & 96.4 & 85.9 & 81.1 & 91.3 & 89.6 & 89.9 & 95.9 & 95.0 & 92.6 & 89.8 (±5.75) \\ \cline{2-3} 
 & GOAD~\cite{bergman2020classification} & ResNet-18 & 75.5 & 94.1 & 81.8 & 72.0 & 83.7 & 84.4 & 82.9 & 93.9 & 92.9 & 89.5 & 85.1 (±7.61) \\ \cline{2-3} 
 & DROC~\cite{sohn2020learning} & ResNet-18 & 90.9 & 98.9 & 88.0 & 83.1 & 89.9 & 90.3 & 93.5 & 98.2 & 96.5 & 95.2 & 92.5 (±4.95) \\ \cline{2-3} 
 & CSI~\cite{tack2020csi} & ResNet-18 & 89.9 & \textbf{99.1} & \textbf{93.1} & 86.4 & 93.9 & \underline{93.2} & 95.1 & \textbf{98.7} & \textbf{97.9} & 95.5 & 94.3 (±3.97) \\ \cline{2-3}
 & FLOW$^{\dag}$~\cite{schmier2022anomaly} & ResNet-50 & 96.1 & 97.5 & 92.6 & 89.8 & 93.3 & \textbf{95.7} & \textbf{98.0} & 94.7 & \underline{97.8} & \underline{96.6} & 95.2 (±2.64) \\
 \hline
\rowcolor{Gray} {\begin{tabular}[c]{@{}c@{}}Federated \\ Learning\end{tabular}} & Ours & ResNet-18 & \textbf{96.4} & \underline{97.9} & \underline{92.7} & \textbf{90.9} & \textbf{95.8} & 92.7 & \underline{96.4} & \underline{96.4} & 97.6 & \textbf{96.7} & \textbf{95.3 (±2.38)} \\ \hline
\end{tabular}
}
\end{center}
\vspace{-2.0em}
\caption{AUROC of various centralized and decentralized detection methods on CIFAR-10 for one-class classification. We present the AUROC of each class and the mean and standard deviation (std) of AUROC for all classes. $^{\dag}$, \textbf{bold}, and \underline{underline} denote the usage of a pre-trained model learned on ImageNet-1M, the best results, and the second results, respectively.}
\vspace{-1.0em}
\label{tab:table3}
\end{table*}

\paragraph{Image and tabular benchmarks}
For the image benchmarks, we present the results on MNIST and CIFAR-10 datasets for OCC in \textbf{extreme} non-i.i.d.($\alpha=0.0$) data setting of FL. Table~\ref{tab:table1} shows the outstanding result of not only the significantly improved performance but also the effective communication cost on FL methods. We found that our proposed method exploits the off-the-shelf model and dataset to compensate the shortage of the local data. For updating a global model every rounds, naive FL methods~\cite{yu2020federated,hosseini2021federated,parkfederated} minimize the distance among the latent variables of the local data and map the distribution of the local data into one secure code designated. Those methods demand the inordinate communication cost and show the limited performance due to the deficiency of the local data on each device and the entangled secure code on latent space. Furthermore, we compare our approach with various centralized methods as depicted in Table~\ref{tab:table2}. Our method outperforms ResNet-18 based methods on CIFAR-10 and ImageNet-30. In case of CIFAR-100\textsuperscript{\ddag}, our approach places the second-rank performance comparing with FLOW~\cite{schmier2022anomaly} leveraging the feature extraction of a pre-trained model learned on ImageNet-1M. Note that our proposed method is the first rank among ResNet-18 based methods having no additional training datasets and limited computing resources. Among FL methods, our proposed method accomplishes the consistently improved performance on the image benchmarks.

\begin{figure}
\centering
    \includegraphics[width=0.7\columnwidth]{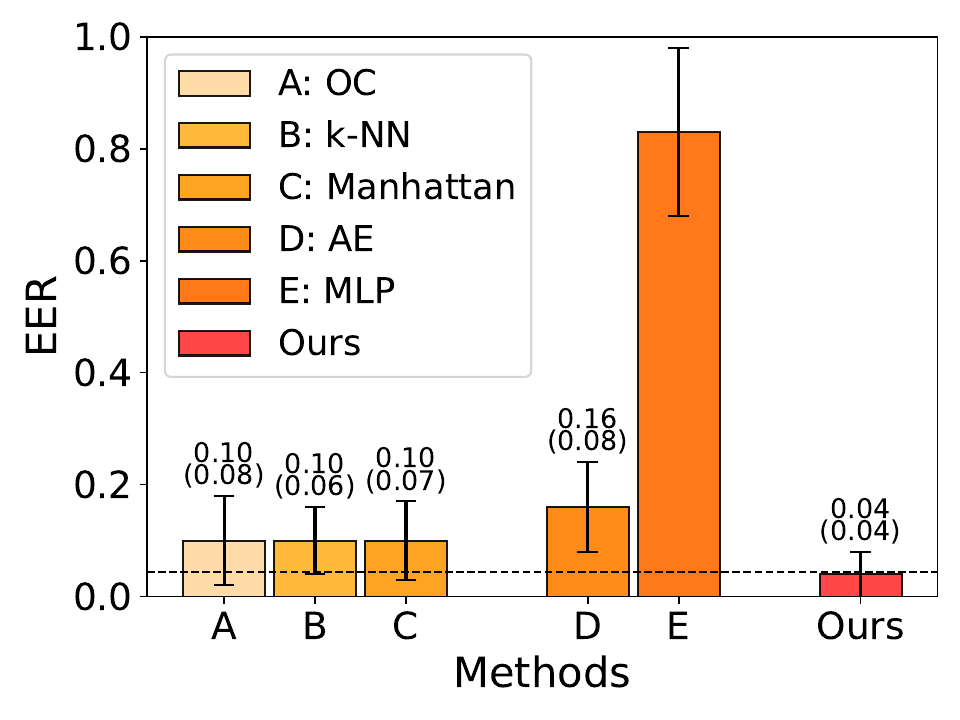}
    \vspace{-1.0em}
    \caption{Comparison to previous methods of one-class classification on the Keystroke-Dynamics dataset. OC, Manhattan, AE, and MLP denote Outlier Count, Euclidean detector with Manhattan distance, Auto-Encoder, and 4-multi layer perception, respectively, from ~\cite{killourhy2009comparing}. $\bottop$ denotes a standard deviation.}
    \vspace{-0.5em}
\label{fig:fig_keystroke}
\end{figure}

For the tabular benchmark, we make the comparisons to the centralized methods on the Keystroke-Dynamics dataset for one-class classification as shown in Fig.~\ref{fig:fig_keystroke}. As the results, we found that our proposed method reduces EER more than two times comparing to prior methods. Training a foundation model as the global model is essential for the OCC task instead of the various architectures and distance metrics to achieve notably improved performance.

\begin{figure}
\centering
    \includegraphics[width=0.7\columnwidth]{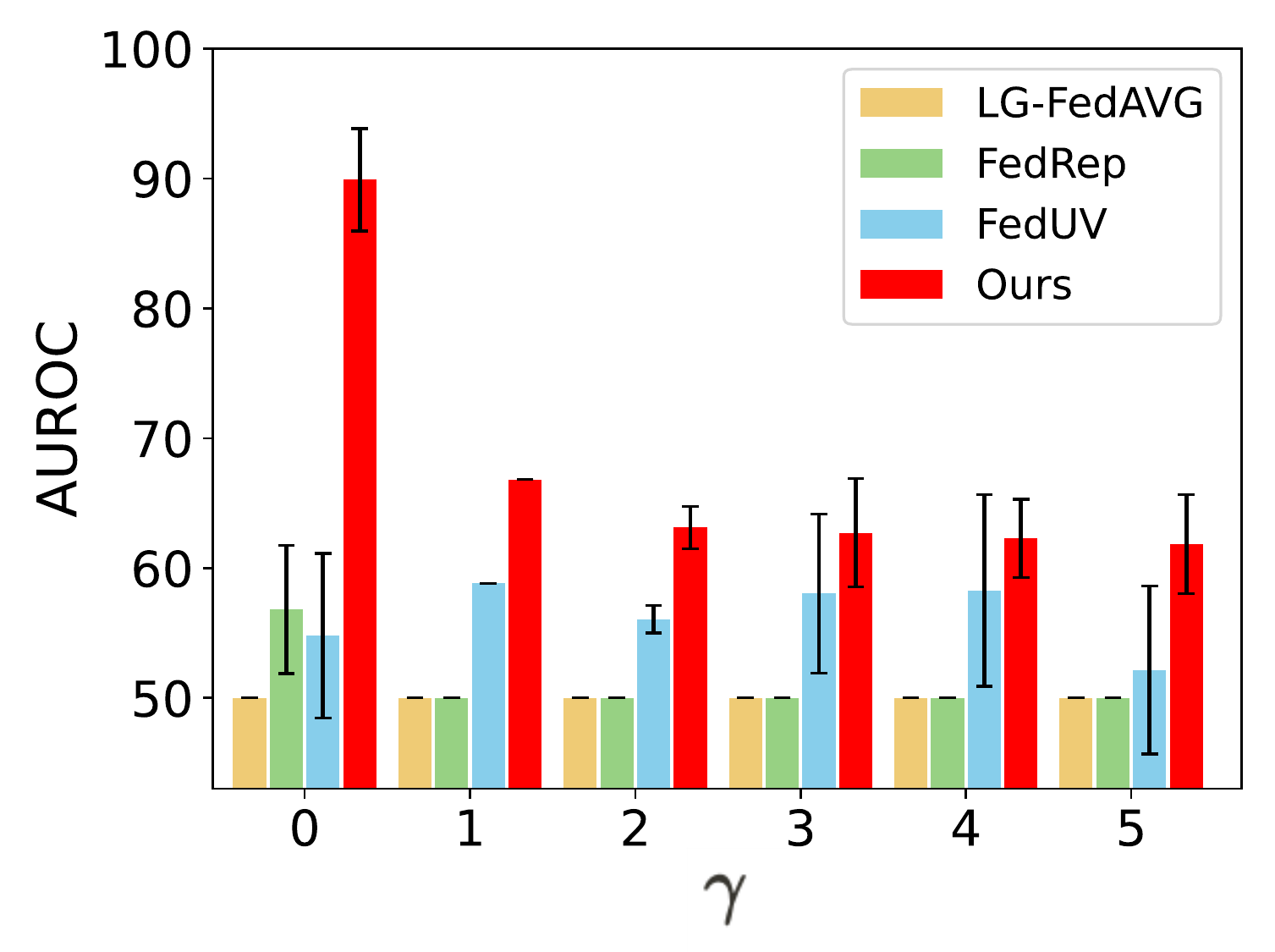}
    \vspace{-1.0em}
    \caption{Analysis the scalability of our global model learned by ProtoFL through new clients on CIFAR-100\textsuperscript{\ddag} dataset. Our approach modestly overcomes the other methods~\cite{collins2021exploiting,hosseini2021federated,liang2020think} in FL with \textbf{extreme} non-i.i.d.($\alpha=0.0$) data. $\gamma$ and $\bottop$ denote the number of new joined clients and a standard deviation each.}
    \vspace{-0.5em}
\label{fig:fig_scalable_representation}
\end{figure}

\vspace{-0.5em}
\paragraph{Centralized vs Federated learning}
As shown in Table~\ref{tab:table3}, we compare our proposed method with the centralized learning based methods because there is no previous FL method studying sufficiently analysis results for CIFAR-10. Our proposed approach outperforms the previous methods in terms of the overall mean and standard deviation because we collaboratively train our global model to represent all classes by the distributively-learned local models, and then we apply our non-biased global model to transform raw data into latent spaces for learning a local one-class classifier in each client as shown in Fig.~\ref{fig:fig_scalable_representation}. In case of each class, this method shows the first and second-placed performance in 8 out of 10, whereas FLOW~\cite{schmier2022anomaly} and CSI~\cite{tack2020csi} show the first and second-placed performance in 4 and 5 out of 10, which indicates that our collaboratively learned global model avoids over-fitting problem for each client.

\vspace{-0.5em}
\paragraph{Scalability of representation}
In this setup on CIFAR-100\textsuperscript{\ddag}, let $\gamma$ indicate the number of new joined clients for one-class classification, and $20-\gamma$ represent the number of clients who participating in FL. To optimize the local classifier parameters of the new client, the new client downloads the global model as described in Algorithm~\ref{alg:cap2}. Note that we are the first to consider the effects of joining new clients. As presented in Fig.~\ref{fig:fig_scalable_representation}, we demonstrate the effectiveness for scalability of the learned representation as our global model through experiments and comparisons. Although new clients have increased, all federated learning methods suffer from degraded performances. However, our proposed method shows more scalability with respect to a new joiner than the others~\cite{hosseini2021federated, collins2021exploiting, liang2020think}. In addition, FedRep~\cite{collins2021exploiting} and LG-FedAvg~\cite{liang2020think} are meaningless to support new clients. Thus, our method presents an novel property validating new joiners performance in FL.

\begin{table}[]
\centering
\resizebox{\columnwidth}{!}{
\begin{tabular}{|c|l|c|c|c||c|}
\hline
Phase 1 & \multicolumn{1}{c|}{Phase 2} & CIFAR-10 & CIFAR-100\textsuperscript{\ddag} & ImageNet-30 & Mean \\ \hline
\multirow{4}{*}{\begin{tabular}[c]{@{}c@{}}Ours \\ (ProtoFL)\end{tabular}} & KDE & 94.5 & 88.9 & 94.0 & 92.3 \\ \cline{2-6} 
 & GDE & 94.3 & 88.6 & 92.8 & 91.9 \\ \cline{2-6} 
 & OC-SVM & 97.1 & 86.9 & 90.8 & 91.6 \\ \cline{2-6}
 & \cellcolor[gray]{0.9}Ours (OC-NF) & \cellcolor[gray]{0.9}\textbf{95.3} & \cellcolor[gray]{0.9}\textbf{89.9} & \cellcolor[gray]{0.9}\textbf{95.4} & \cellcolor[gray]{0.9}\textbf{93.5} \\ \hline
\end{tabular}
}
\vspace{-1.0em}
\caption{Ablation study of various classifiers on CIFAR-10, CIFAR-100\textsuperscript{\ddag}, and ImageNet-30 with our representative global model. The performance is evaluated by AUROC.}
\label{tab:table4}
\end{table}

\begin{table}[]
\centering
\resizebox{\columnwidth}{!}{
\begin{tabular}{|l|ccc|cc|c|c|}
\hline
\multicolumn{1}{|c|}{} & \multicolumn{2}{c|}{Phase 1} & \multicolumn{2}{c|}{Phase 2} & CIFAR-100\textsuperscript{\ddag} & ImageNet-30 \\ \cline{2-7} 
\multicolumn{1}{|c|}{\multirow{-2}{*}{Method}} & \multicolumn{1}{c|}{$L^{\theta}_{pd}$}& $L^{\theta}_{p}$ & \multicolumn{1}{|c|}{$L^{\psi}_{mle}$} & \multicolumn{1}{|c|}{$L^{\psi}_{reg}$} & AUROC & AUROC \\ \hline
Ours w/o $L^{\theta}_{pd}, L^{\psi}_{reg}$ & \multicolumn{1}{c|}{} & \checkmark & \multicolumn{1}{|c|}{\checkmark} & \multicolumn{1}{c|}{} & 48.9 & 50.2 \\ \hline
Ours w/o $L^{\theta}_{p}, L^{\psi}_{reg}$ & \multicolumn{1}{c|}{\checkmark} &  & \multicolumn{1}{|c|}{\checkmark} & \multicolumn{1}{c|}{} & 87.0 & 91.1 \\ \hline
Ours w/o $L^{\psi}_{reg}$ & \multicolumn{1}{c|}{\checkmark} & \checkmark & \multicolumn{1}{|c|}{\checkmark} & \multicolumn{1}{c|}{} & 89.6 & 94.4 \\ \hline
\rowcolor{Gray}Ours & \multicolumn{1}{c|}{\checkmark} & \checkmark & \multicolumn{1}{|c|}{\checkmark} & \multicolumn{1}{c|}{\checkmark} & \textbf{89.9} & \textbf{95.4} \\ \hline
\end{tabular}
}
\vspace{-1.0em}
\caption{Ablation study on CIFAR-100\textsuperscript{\ddag} and ImageNet-30 to evaluate each component in the individual phase.}
\vspace{-1.0em}
\label{tab:table5}
\end{table}

\subsection{Ablation Studies}
We conducted ablation studies on CIFAR-10/100\textsuperscript{\ddag} and ImageNet-30 (a)to verify the phase 1 and the phase 2, (b)to explore the contribution of each component in our method, and (c)to investigate the effectiveness of the off-the-shelf model in respect of the various off-the-shelf datasets.

\vspace{-0.3em}
\paragraph{Analyses of our global- and classifier- model}
With fixed our global model learned by joined clients in advance, we compared our proposed local one-class classifier via normalizing flows (OC-NF) with the prior one-class classifiers for the usability of the global model. As shown in Table~\ref{tab:table4}, we observed the superiority of the global model and the local classifier either each or both comparing to the previous detectors~\cite{kde,scholkopf1999support} non-parametric kernel density estimation, parametric gaussian density estimation, and one-class SVM (KDE, GDE, and OC-SVM).

\begin{figure}[]
    \centering
    \includegraphics[width=\columnwidth]{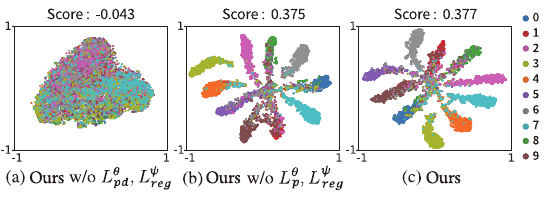}
    \vspace{-2.0em}
    \caption{Visualization of latent-features using t-sne for our global model learned (\textit{Phase 1}) with either each component or all. \emph{Score} in each figure denote the silhouette values~\cite{rousseeuw1987silhouettes}.}
\label{fig:fig_performance}
\vspace{-1.0em}
\end{figure}

\vspace{-0.3em}
\paragraph{Analyses of each component}
To analyze the effect of our proposed objective for each phase, we separated the representation objective of the phase 1 into distilling loss $\mathcal{L}^{\theta}_{pd}$ and similarity loss $\mathcal{L}^{\theta}_{p}$, and the classification objective of the phase 2 into maximum likelihood $\mathcal{L}^{\psi}_{mle}$ and regularizer $\mathcal{L}^{\psi}_{reg}$ to show the importance in Table ~\ref{tab:table5}. Thus, we confirmed that each of them is the essential terms for the improving performance, and also observed the best results when all of them were used on CIFAR-100\textsuperscript{\ddag} and ImageNet-30. In case of ImageNet-30, the improved performance indicates the effectiveness of this regularizer. As shown in Fig.~\ref{fig:fig_performance}, we found that our representative global model, trained without this distilling loss $\mathcal{L}^{\theta}_{pd}$, leads the drastically impoverished representation.

\begin{figure}[]
\begin{subfigure}{.23\textwidth}
  \centering
  \includegraphics[width=\linewidth]{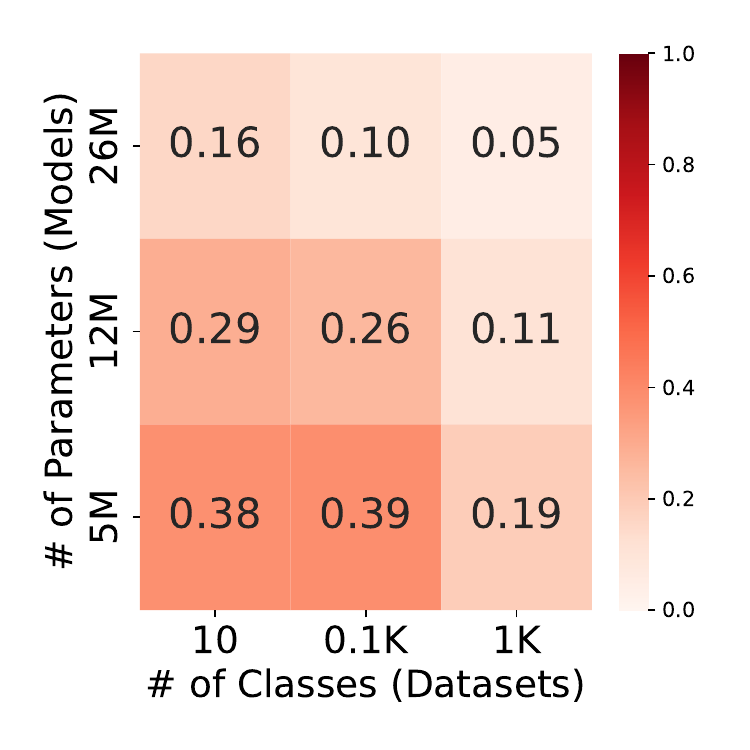}
  \label{fig:sub-domain-first}
\end{subfigure}
\begin{subfigure}{.23\textwidth}
  \centering
  \includegraphics[width=\linewidth]{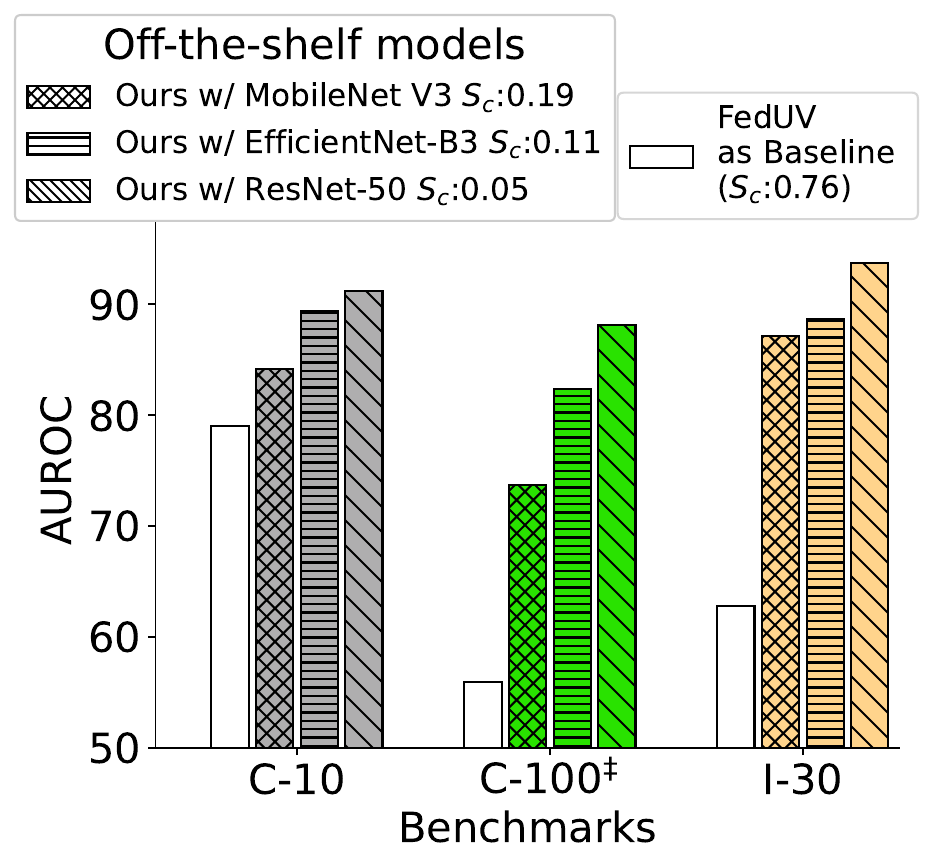}
  \label{fig:sub-domain-second}
\end{subfigure}
\vspace{-2.0em}
\caption{Left : Each score denotes cosine similarity $\mathcal{S}_{c}$ between the various off-the-shelf datasets and models. Right : Performance comparison among the various global models whose off-the-shelf model is different. C and I denote CIFAR and ImageNet.}
\vspace{-1.0em}
\label{fig:fig_domain}
\end{figure}

\vspace{-0.3em}
\paragraph{Analyses of distilling representation}
With the FL setting, we employ cosine similarity $\mathcal{S}_{c}$ as a criterion for selecting the off-the-shelf models as shown in left of Fig.~\ref{fig:fig_domain}, and evaluated the performance based on various $\mathcal{S}_{c}$ on all the benchmarks as shown in right of Fig.~\ref{fig:fig_domain}. Our proposed ProtoFL achieves the significantly improved performance comparing to FedUV since the ProtoFL is robust to the increased clients for one-class classification, and the results show that lower $\mathcal{S}_{c}$ indicate more discriminative features. Note, distributing the prototypical representation in secure is important to overcome the shortage of training data, limited round communications, and greater number of clients.

\vspace{-0.3em}
\section{Conclusion}
We proposed ProtoFL, a method for achieving effective round cost and scalability representation in extreme non-i.i.d. data based FL. By utilizing an off-the-shelf model and dataset to distribute prototypical independent representations, we were able to learn a global model with all joined clients and optimize the flow-based classifier of each client. Our proposed method outperforms FL methods with efficient communication cost and presents a novel property of scalability of representation, validating new joiners' performance in FL. We believe that our method and experimental details could be adapted to handle further problems in both federated and centralized learning based one-class classification.

\vspace{-0.3em}
\section{Acknowledgements}
\label{sec:acknowledgements}
\vspace{-0.8em}

This work was supported by KakaoBank Corp., and Institute of Information \& communications Technology Planning \& Evaluation (IITP) grant funded by the Korea government (MSIT) (No. 2022-0-00320, Artificial intelligence research about cross-modal dialogue modeling for one-on-one multi-modal interactions). In particular, we would like to thank the designer Eunha Lim (\textit{milkywaymind@naver.com}) for well-presented Figures ~\ref{fig:fig_title}, ~\ref{fig:fig_main}, and ~\ref{fig:fig_compare}.

{\small
\bibliographystyle{ieee_fullname}
\bibliography{egbib}

\begin{thebibliography}{10}\itemsep=-1pt

\bibitem{CHO2022108703}
Unsupervised video anomaly detection via normalizing flows with implicit latent
  features.
\newblock {\em Pattern Recognition}, 129:108703, 2022.

\bibitem{fedface}
Divyansh Aggarwal, Jiayu Zhou, and Anil~K. Jain.
\newblock Fedface: Collaborative learning of face recognition model.
\newblock {\em In IJCB}, 2021.

\bibitem{bergman2020classification}
Liron Bergman and Yedid Hoshen.
\newblock Classification-based anomaly detection for general data.
\newblock {\em arXiv preprint arXiv:2005.02359}, 2020.

\bibitem{kde}
Markus~M Breunig, Hans-Peter Kriegel, Raymond~T Ng, and J{\"o}rg Sander.
\newblock Lof: identifying density-based local outliers.
\newblock In {\em ACM sigmod record}, number~2. ACM, 2000.

\bibitem{chen2020simple}
Ting Chen, Simon Kornblith, Mohammad Norouzi, and Geoffrey Hinton.
\newblock A simple framework for contrastive learning of visual
  representations.
\newblock In {\em International conference on machine learning}, pages
  1597--1607. PMLR, 2020.

\bibitem{collins2021exploiting}
Liam Collins, Hamed Hassani, Aryan Mokhtari, and Sanjay Shakkottai.
\newblock Exploiting shared representations for personalized federated
  learning.
\newblock In {\em International Conference on Machine Learning}, pages
  2089--2099. PMLR, 2021.

\bibitem{Deng_2019_CVPR}
Jiankang Deng, Jia Guo, Niannan Xue, and Stefanos Zafeiriou.
\newblock Arcface: Additive angular margin loss for deep face recognition.
\newblock In {\em CVPR}, June 2019.

\bibitem{dinh2017density}
Laurent Dinh, Jascha Sohl-Dickstein, and Samy Bengio.
\newblock Density estimation using real {NVP}.
\newblock In {\em International Conference on Learning Representations}, 2017.

\bibitem{SphereFed}
Xin Dong, Sai~Qian Zhang, Ang Li, and H.~T. Kung.
\newblock Spherefed: Hyperspherical federated learning.
\newblock In {\em ECCV}, 2022.

\bibitem{golan2018deep}
Izhak Golan and Ran El-Yaniv.
\newblock Deep anomaly detection using geometric transformations.
\newblock {\em Advances in neural information processing systems}, 31, 2018.

\bibitem{gudovskiy2022cflow}
Denis Gudovskiy, Shun Ishizaka, and Kazuki Kozuka.
\newblock Cflow-ad: Real-time unsupervised anomaly detection with localization
  via conditional normalizing flows.
\newblock In {\em WACV}, pages 98--107, 2022.

\bibitem{guo2016ms}
Yandong Guo, Lei Zhang, Yuxiao Hu, Xiaodong He, and Jianfeng Gao.
\newblock Ms-celeb-1m: A dataset and benchmark for large-scale face
  recognition.
\newblock In {\em ECCV}, pages 87--102. Springer, 2016.

\bibitem{han2022fedx}
Sungwon Han, Sungwon Park, Fangzhao Wu, Sundong Kim, Chuhan Wu, Xing Xie, and
  Meeyoung Cha.
\newblock {FedX: Unsupervised Federated Learning with Cross Knowledge
  Distillation}.
\newblock In {\em ECCV}, 2022.

\bibitem{he2016deep}
Kaiming He, Xiangyu Zhang, Shaoqing Ren, and Jian Sun.
\newblock Deep residual learning for image recognition.
\newblock In {\em CVPR}, pages 770--778, 2016.

\bibitem{hendrycks2019oe}
Dan Hendrycks, Mantas Mazeika, and Thomas Dietterich.
\newblock Deep anomaly detection with outlier exposure.
\newblock {\em Proceedings of the International Conference on Learning
  Representations}, 2019.

\bibitem{hendrycks2019using}
Dan Hendrycks, Mantas Mazeika, Saurav Kadavath, and Dawn Song.
\newblock Using self-supervised learning can improve model robustness and
  uncertainty.
\newblock {\em Advances in neural information processing systems}, 32, 2019.

\bibitem{hosseini2021federated}
Hossein Hosseini, Hyunsin Park, Sungrack Yun, Christos Louizos, Joseph Soriaga,
  and Max Welling.
\newblock Federated learning of user verification models without sharing
  embeddings.
\newblock In {\em International Conference on Machine Learning}, pages
  4328--4336. PMLR, 2021.

\bibitem{noniid}
Tzu{-}Ming~Harry Hsu, Hang Qi, and Matthew Brown.
\newblock Measuring the effects of non-identical data distribution for
  federated visual classification.
\newblock 2019.

\bibitem{killourhy2009comparing}
Kevin~S Killourhy and Roy~A Maxion.
\newblock Comparing anomaly-detection algorithms for keystroke dynamics.
\newblock In {\em 2009 IEEE/IFIP International Conference on Dependable Systems
  \& Networks}, pages 125--134. IEEE, 2009.

\bibitem{NEURIPS2020_ecb9fe2f}
Polina Kirichenko, Pavel Izmailov, and Andrew~G Wilson.
\newblock Why normalizing flows fail to detect out-of-distribution data.
\newblock In H. Larochelle, M. Ranzato, R. Hadsell, M.F. Balcan, and H. Lin,
  editors, {\em Advances in Neural Information Processing Systems}, volume~33,
  pages 20578--20589. Curran Associates, Inc., 2020.

\bibitem{krizhevsky2009learning}
Alex Krizhevsky, Geoffrey Hinton, et~al.
\newblock Learning multiple layers of features from tiny images.
\newblock 2009.

\bibitem{fedfv}
Liu L, Zhang Y, Gao H, and et al.
\newblock Fedfv: federated face verification via equivalent class embeddings.

\bibitem{lecun2010mnist}
Yann LeCun, Corinna Cortes, and CJ Burges.
\newblock Mnist handwritten digit database. att labs, 2010.

\bibitem{liang2020think}
Paul~Pu Liang, Terrance Liu, Liu Ziyin, Nicholas~B Allen, Randy~P Auerbach,
  David Brent, Ruslan Salakhutdinov, and Louis-Philippe Morency.
\newblock Think locally, act globally: Federated learning with local and global
  representations.
\newblock {\em arXiv preprint arXiv:2001.01523}, 2020.

\bibitem{fedfr}
Chih{-}Ting Liu, Chien{-}Yi Wang, Shao{-}Yi Chien, and Shang{-}Hong Lai.
\newblock Fedfr: Joint optimization federated framework for generic and
  personalized face recognition.
\newblock abs/2112.12496.

\bibitem{liu2019variance}
Liyuan Liu, Haoming Jiang, Pengcheng He, Weizhu Chen, Xiaodong Liu, Jianfeng
  Gao, and Jiawei Han.
\newblock On the variance of the adaptive learning rate and beyond.
\newblock {\em arXiv preprint arXiv:1908.03265}, 2019.

\bibitem{liznerski2021explainable}
Philipp Liznerski, Lukas Ruff, Robert~A. Vandermeulen, Billy~Joe Franks, Marius
  Kloft, and Klaus~Robert Muller.
\newblock Explainable deep one-class classification.
\newblock In {\em International Conference on Learning Representations}, 2021.

\bibitem{liznerski2022exposing}
Philipp Liznerski, Lukas Ruff, Robert~A. Vandermeulen, Billy~Joe Franks,
  Klaus~Robert Muller, and Marius Kloft.
\newblock Exposing outlier exposure: What can be learned from few, one, and
  zero outlier images.
\newblock {\em In TMLR}, 2022.

\bibitem{Orchestra}
Ekdeep~Singh Lubana, Chi~Ian Tang, Fahim Kawsar, Robert~P. Dick, and Akhil
  Mathur.
\newblock Orchestra: Unsupervised federated learning via globally consistent
  clustering, 2022.

\bibitem{mcmahan2017communication}
Brendan McMahan, Eider Moore, Daniel Ramage, Seth Hampson, and Blaise~Aguera y
  Arcas.
\newblock Communication-efficient learning of deep networks from decentralized
  data.
\newblock In {\em Artificial intelligence and statistics}, pages 1273--1282.
  PMLR, 2017.

\bibitem{osti_6755553}
M~M Moya, M~W Koch, and L~D Hostetler.
\newblock One-class classifier networks for target recognition applications.
\newblock 1 1993.

\bibitem{fedaa}
Poojan Oza and Vishal~M. Patel.
\newblock Federated learning-based active authentication on mobile devices.
\newblock 2021.

\bibitem{JMLR:v22:19-1028}
George Papamakarios, Eric Nalisnick, Danilo~Jimenez Rezende, Shakir Mohamed,
  and Balaji Lakshminarayanan.
\newblock Normalizing flows for probabilistic modeling and inference.
\newblock {\em In JMLR}, 2021.

\bibitem{parkfederated}
Hyunsin Park, Hossein Hosseini, and Sungrack Yun.
\newblock Federated learning with metric loss.

\bibitem{robbins1951stochastic}
Herbert Robbins and Sutton Monro.
\newblock A stochastic approximation method.
\newblock {\em The annals of mathematical statistics}, pages 400--407, 1951.

\bibitem{rousseeuw1987silhouettes}
Peter~J Rousseeuw.
\newblock Silhouettes: a graphical aid to the interpretation and validation of
  cluster analysis.
\newblock {\em Journal of computational and applied mathematics}, 20:53--65,
  1987.

\bibitem{ruff2018deep}
Lukas Ruff, Robert Vandermeulen, Nico Goernitz, Lucas Deecke, Shoaib~Ahmed
  Siddiqui, Alexander Binder, Emmanuel M{\"u}ller, and Marius Kloft.
\newblock Deep one-class classification.
\newblock In {\em International conference on machine learning}, pages
  4393--4402. PMLR, 2018.

\bibitem{Ruff2020Deep}
Lukas Ruff, Robert~A. Vandermeulen, Nico Görnitz, Alexander Binder, Emmanuel
  Müller, Klaus-Robert Müller, and Marius Kloft.
\newblock Deep semi-supervised anomaly detection.
\newblock In {\em International Conference on Learning Representations}, 2020.

\bibitem{nfad}
Artem Ryzhikov, Maxim Borisyak, Andrey Ustyuzhanin, and Denis Derkach.
\newblock Nfad: fixing anomaly detection using normalizing flows.
\newblock {\em PeerJ Computer Science}, 2021.

\bibitem{schmier2022anomaly}
Robert Schmier, Ullrich K{\"o}the, and Christoph-Nikolas Straehle.
\newblock Anomaly detection using contrastive normalizing flows.
\newblock {\em arXiv preprint arXiv:2208.14024}, 2022.

\bibitem{scholkopf1999support}
Bernhard Sch{\"o}lkopf, Robert~C Williamson, Alex Smola, John Shawe-Taylor, and
  John Platt.
\newblock Support vector method for novelty detection.
\newblock {\em Advances in neural information processing systems}, 12, 1999.

\bibitem{sohn2020learning}
Kihyuk Sohn, Chun-Liang Li, Jinsung Yoon, Minho Jin, and Tomas Pfister.
\newblock Learning and evaluating representations for deep one-class
  classification.
\newblock {\em arXiv preprint arXiv:2011.02578}, 2020.

\bibitem{tack2020csi}
Jihoon Tack, Sangwoo Mo, Jongheon Jeong, and Jinwoo Shin.
\newblock Csi: Novelty detection via contrastive learning on distributionally
  shifted instances.
\newblock {\em Advances in neural information processing systems},
  33:11839--11852, 2020.

\bibitem{truong2021privacy}
Nguyen Truong, Kai Sun, Siyao Wang, Florian Guitton, and YiKe Guo.
\newblock Privacy preservation in federated learning: An insightful survey from
  the gdpr perspective.
\newblock {\em Computers \& Security}, 110:102402, 2021.

\bibitem{yu2020federated}
Felix Yu, Ankit~Singh Rawat, Aditya Menon, and Sanjiv Kumar.
\newblock Federated learning with only positive labels.
\newblock In {\em International Conference on Machine Learning}, pages
  10946--10956. PMLR, 2020.

\end{thebibliography}
}
\end{document}